\documentclass[letterpaper, 10 pt, conference]{ieeeconf}  % Comment this line out if you need a4paper

\IEEEoverridecommandlockouts                              % This command is only needed if 
\usepackage{graphicx}
\usepackage{hyperref}
\usepackage{xcolor}
\usepackage{amsmath} % assumes amsmath package installed
\usepackage{amssymb}  % assumes amsmath package installed
\begin{document}
\title{\LARGE \bf
Intention estimation from gaze and motion features for human-robot shared-control object manipulation
}

\author{Anna Belardinelli$^{1}$, Anirudh Reddy Kondapally$^{2}$, Dirk Ruiken${^1}$, Daniel Tanneberg${^1}$, and Tomoki Watabe$^{2}$ %<-this % stops a space
%\thanks{*This work was not supported by any organization}% <-this % stops a space
\thanks{$^{1}$ AB, DR, and DT are with the Honda Research Institute EU, Offenbach, Germany
        {\tt\small \{anna.belardinelli, dirk.ruiken, daniel.tanneberg\}@honda-ri.de}}%
\thanks{$^{2}$ ARK and TW are with the Honda R\&D Co., Ltd. Innovative Research Excellence, Japan {\tt\small \{anirudh\_kondapally-reddy, tomoki\_watabe\}@jp.honda}}%
}

\maketitle
\thispagestyle{empty}
\pagestyle{empty}

%%%%%%%%%%%%%%%%%%%%%%%%%%%%%%%%%%%%%%%%%%%%%%%%%%%%%%%%%%%%%%%%%%%%%%%%%%%%%%%%
\begin{abstract}

Shared control can help in teleoperated object
manipulation by assisting with the execution of the user’s
intention. To this end, robust and prompt intention estimation
is needed, which relies on behavioral observations. Here, an
intention estimation framework is presented, which uses natural gaze and motion features to predict the current action and the target object. The system is trained and tested in a simulated environment with pick and place sequences produced in a relatively cluttered scene and with both hands, with possible hand-over to the other hand. Validation is conducted across different users and hands, achieving good accuracy and earliness of prediction. An analysis of the predictive power of single features shows the predominance of the grasping trigger and the gaze features in the early identification of the current action. In the current framework, the same probabilistic model can be used for the two hands working in parallel and independently, while a rule-based model is proposed to identify the resulting bimanual action. Finally, limitations and perspectives of this approach to more complex, full-bimanual manipulations are discussed.

\end{abstract}

%%%%%%%%%%%%%%%%%%%%%%%%%%%%%%%%%%%%%%%%%%%%%%%%%%%%%%%%%%%%%%%%%%%%%%%%%%%%%%%%
\section{INTRODUCTION}

The need for dexterous manipulation capabilities in robotic teleoperation has been rising in recent years, especially beyond classical industrial applications, for example in  remote surgery \cite{Choi2018}, assistive robotic limbs for impaired users \cite{Herlant2016}, spatial missions \cite{Artigas2016}, or nuclear waste disposal \cite{Petereit2019}. As the recent pandemic has shown, whenever a flexible and skilled manual action is required but human physical presence is not possible, teleoperation could provide a means to remedy the situation, may it be with telepresence for helping distant relatives, in healthcare to safely provide medical assistance to contagious patients, in industrial productions requiring sterile environments, or in assistive applications restoring arm mobility to impaired users. In such cases, even when no specific tool is used and motion input control could be in principle delivered by natural limb movements (unlike in some assistive scenarios using joysticks or brain-computer interfaces), it might still be difficult to operate a robot with little to no training.
\begin{figure}[t]
    \centering
\includegraphics[width=0.45\textwidth]{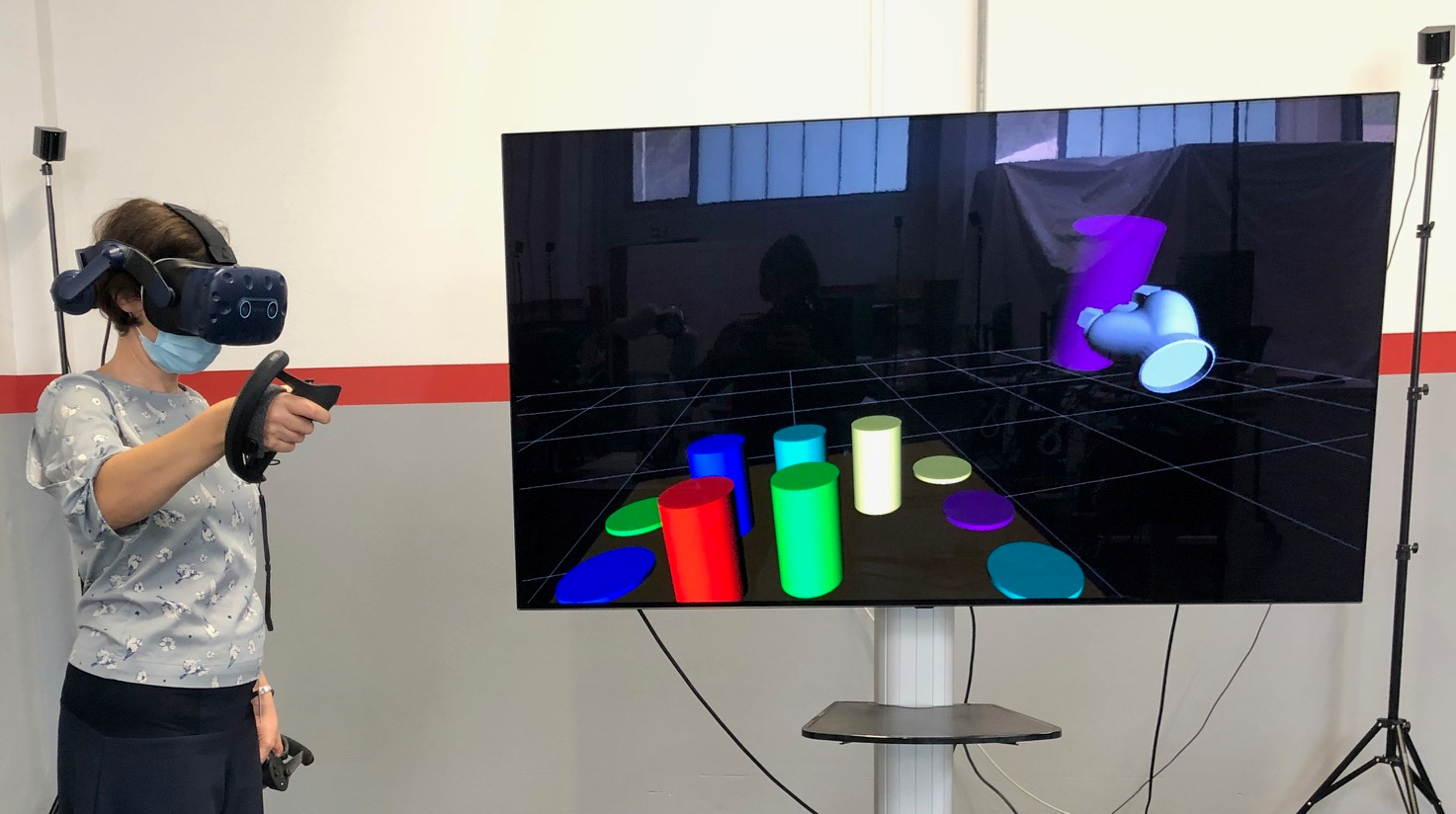}
\caption{Experimental setup for the data collection and model testing. A robotic hand is operated in virtual reality to execute pick and place tasks. The participant's head and hands movements are tracked via the Vive Lighthouse system, the eyes are tracked inside the Vive headset. The screen on the background visualizes the scene as perceived by the participant's point of view.}
\label{fig:layout}
\end{figure}
Since perception and action are in this case both mediated by technical systems, they are also possibly affected by delays. Further, usually no or limited haptic feedback is available. Therefore, these applications would also benefit from shared control manipulation \cite{Griffin2005, Rakita2019}. In such cases, the user could interact with objects more naturally, exploiting their natural eye-hand coordination, while the robotic systems translates their intentions into the real scene and relieves them of low-level kinematic control. We recently proposed an intention estimation framework relying on gaze features to infer pick and place intentions and related targets in a rather simplified scene with just 4 objects and 6 intentions \cite{Fuchs2021}. Here, that scenario is extended to work with a more cluttered scene with partially occluded objects that can be grasped with either hand. Hence, here the following contributions are made:
\begin{itemize}
\item A new dataset was collected in virtual reality, with multiple objects purposely placed close together to create visual occlusions. This configuration elicits complex gaze behaviors and different grasping movements, depending on the distractors around the target object and on the used hand. 
\item The second hand was also introduced in the virtual environment to evaluate the generalization of the intention recognition model across effectors.
\item The feature set for the model is expanded to take into account the current motion trajectory of the arm, which can be suggestive of the target of the intended interaction. 
\item We propose that in this way a single-hand framework can be straightforwardly extended to bimanual manipulations, combining intentions from two hands.
\end{itemize}

The experimental setup for data collection and the feature extraction procedures are presented in Section \ref{sec:exp}. The system was tested on different users and across actions performed with different hands, to demonstrate its robustness and to evaluate the predictive capability of different feature combinations. The new model and the test results are presented in Sections \ref{sec:model}, \ref{sec:bim}, and \ref{sec:results}.  In section \ref{sec:conclusions} we conclude with a discussion on current developments and possible future extensions of the bimanual architecture.

\section{RELATED WORK}

Within shared autonomy, first approaches to intention estimation and several recent ones  have made use of the user control input in driving the robotic movement as behavioral cue for prediction \cite{Hauser2013,Dragan2013,Javdani2015, Aarno08,Tanwani2017}. This could be given via more or less intuitive input modalities, such as mouse pointers \cite{Hauser2013}, joysticks \cite{Javdani2015}, another robot arm \cite{Tanwani2017}, or the user's movements as tracked by a camera \cite{Dragan2013}. Most of these works rely on motion features such as effector pose, velocity, arm joints, or whole gestemes, to predict a distribution over the different action targets.

On the other hand, a number of more recent approaches have exploited last decades' advances in eye tracking capabilities to infer intentions from head-mounted eye trackers. A large body of literature, indeed, has shown how during daily manipulations the eyes anticipate the hand, landing on the targeted object about 0.5 s before making contact with it \cite{Hayhoe2003,Johansson2001,Belardinelli2016}.
Many gaze-based frameworks are used in assistive applications where natural movements might not be an option and control input is provided by joystick or rehabilitation robotic arms \cite{Shafti2019,Aronson2021}. In these cases, the eyes indeed offer a much earlier and less noisy cue for intention estimation, while the effector movements are controlled in a non-usual fashion.
Considering intention estimation from egocentric vision in natural eye-hand coordination, \cite{Fathaliyan2018} and more recently \cite{Wang2020} proposed a method for intention recognition in manipulation sequences, relying on time series of gazed objects and angular velocities of the gaze. 
Still, in principle, a number of behavioral cues can be used for intention estimation, leveraging the different predictive power. This was already postulated for instance in \cite{Admoni2016} and \cite{Jain2019}, where general frameworks for intent inference from any number of cues were put forward.
In recent years, some studies have looked at the effective benefit of feeding both gaze and motion features into the prediction system. In \cite{Razin2017}, numerous prediction techniques were benchmarked in the context of teleoperating a robotic claw by means of three levers and considering just 'move', 'hold', and 'stop' intentions. In this simplified scenario, motion features such as the claw position and velocity were enough to reach a good prediction accuracy, with gaze position and velocity \textit{per se} also demonstrating a good predictive power but not improving the performance when combined with motion features. Hidden Markov Models were further among the best scoring classification methods, as shown by their large use in different forms in the literature altogether (e.g., \cite{Aarno08,Tanwani2017,Fuchs2021,Aronson2021}). In \cite{Li2020}, yet, more complex manipulations were considered (handover, transfer, usage) and both the eyes and the hands were tracked for intent recognition also from grasp modeling. In this scenario, the authors found that gaze  helped in improving performance in teleoperation in the case of ambiguous grasps. For these reasons, and with the ultimate goal of prospectively moving  to full bimanual intention recognition, the HMM-based framework presented in \cite{Fuchs2021} is extended with motion features, a richer dataset is collected, and relative challenges and improvements are investigated. For congruence with the gaze features, we used motion features that, as the Areas-Of-Interest recalled below, denote not just an effector pose or motion characteristics, but already entail the likelihood of each object being the target of the current movement trajectories, as detailed in the next section.

\section{DATA COLLECTION AND PROCESSING }
\label{sec:exp}
\begin{figure}[bt]
    \centering
\includegraphics[width=0.48\textwidth]{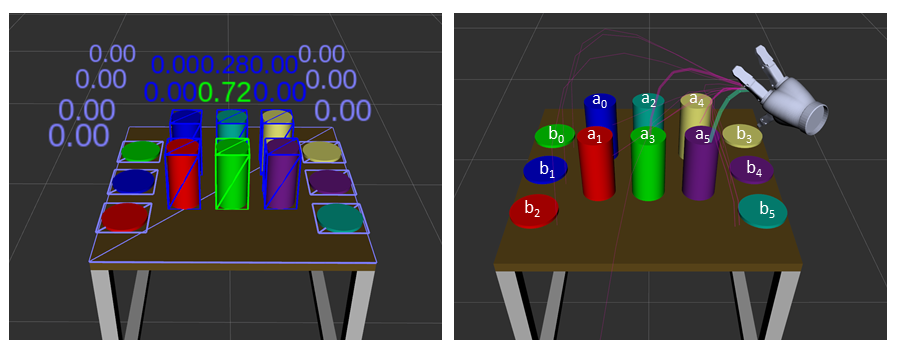}
\caption{Explanatory visualizations of the Areas-Of-Interest (AOIs, left) and Target-Path-Analysis  features (TPAs, right). For the AOIs, the bounding box of the currently gazed object is highlighted in green along with the corresponding normalized likelihood. For the TPAs, the most likely trajectory to each object is shown and the most close to the current predicted hand trajectory is highlighted in green (i.e., to the purple cylinder with approach point on top). The other trajectories are depicted in magenta with line width proportional to their likelihood. In white, also the location labels are depicted: $a_i$ labels refer to 'pickable' cylinders, $b_i$ to placing locations. Features and labels were not visible to the participants performing the tasks. The left hand is not visible in the field of view here, yet a TPA value for it was still computed.}
\label{fig:AOI_TPA_vis}
\end{figure}
\subsection{Experimental setup}
Pick-and-place sequences were collected from human users in a virtual scene, with physics simulated in the Gazebo simulator and rendered to the user via Rviz. This was displayed in the HTC Vive Pro Eye headset, featuring a 1440 x 1600 pixels screen per eye (2880 x 1600 pixels combined, 110$^{\circ}$ Field-Of-View), and a binocular Tobii eyetracker working at 120 Hz with a stated accuracy of 0.5$^{\circ}$ to 1.1$^{\circ}$ (see Fig. \ref{fig:layout} for the physical setup).
The virtual scene consisted of a table with 6 cylinders (20\,cm high and with a radius of 5\,cm) arranged in two lines in front of the participant. The cylinders were spaced 15\,cm apart on the horizontal axis and 10\,cm apart along the vertical axis, on the table plane. The cylinders are conventionally labeled as $\{a0,..,a5\}$, with even numbers for the furthest line and odd numbers for the line closest to the participant (see Fig.~\ref{fig:AOI_TPA_vis}). Further, 6 coasters were presented, aligned along the vertical edges of the table (3 by each side), spaced 20 cm from one another vertically and 66 cm apart horizontally. These discs defined the placing target positions, named as $\{b0,..,b5\}$. For each participant the height of the table was adjusted so that from their egocentric perspective only about the upper half size of the cylinders on the second line was visible. In this way, we wanted to assess the visual behavior in terms of gaze data and the ensuing robustness of the recognition system in case of occluded objects. We decided to work with cylinders  since we were interested here in investigating the gaze and grasping behavior of generic objects in the presence of crowding and occlusions, independently of their identity. The visuomotor behavior with real objects indeed is often influenced by their goal-driven or habitual use \cite{Herbort2011,Belardinelli2016}.\par
In each trial, the participant would first see an occluding screen where the word of the color defining the pick and place tasks was presented. Upon disappearance of this screen (after 3 seconds), they were requested to grasp the cylinder in the given color and place it on the coaster of the same color. Each participant carried out 4 blocks of trials: one with the right hand, one with the left, two with hand-overs (picking up with the right and placing down with the left hand and vice versa). Each block contained 36 trials, i.e., each combination of picking and placing targets. The Vive controllers, tracked by the Vive Lighthouse system, were used to grasp objects. With their movements participants could operate two Kinova robotic grippers and grasp an object by pressing the button on the controller. Only the grippers, i.e. no arms, were visible in the scene and their position and orientation reflected the hand pose of the participants (see \cite{Fuchs2021} for more details on a similar setup). Colors were randomized across both pick and place positions in each trial.  Data were collected from 10 non-naive Honda Research Institute associates (9 males, 1 female) working on a related project, all right-handed.

\subsection{Gaze-based features: Areas-Of-Interest}
Each object in the scene (cylinders, coasters and robot hands) is considered as an Area-Of-Interest (AOI), that is, a stimulus that can be the target of visual attention. As a proxy to measure visual attention on each object, we consider the current gaze position $\bf{g}_t$ at any time as the center of a Gaussian distribution $\mathcal{N}(\bf{g_t},\Sigma)$, with variance defined by a radius of 1$^\circ$ (accounting for eyetracker accuracy and human fovea size \cite{Strasburger2011}).  The surface integral of the density function over the visible faces of each object's bounding box represents the likelihood that this area is the current focus of attention \cite{Fuchs2021}. A depth buffer is used to determine occlusions w.r.t. the participants' point of view and to compute the likelihood only on the portions of the object actually visible to them. The likelihood was normalized across AOIs so to consider a distribution only on those (the table was not relevant to the task and was not considered since it would collect incidental gaze points).

\subsection{Hand motion features: Grasping state and Target Path Analysis}
Besides gaze features, two other types of features were logged in every trial. 
The state of the grasping trigger (as a binary feature) and a similarity measure between the predicted trajectory of the acting hand and an ideal trajectory to each object in the scene (as for the AOIs) and to the other hand (see Fig.~\ref{fig:AOI_TPA_vis}, right). We term this latter Target Path Analysis (TPA) and it is inspired by the prediction of risk collision between two trajectories sketched in \cite{Muehlig2014} and formalized in \cite{Iba2019}. TPA features are computed as follows. 

The current acting hand pose is taken as input along with each object pose. For each object a number of approach points are computed, by considering a location on the object as destination of the hand movement and the direction of approach of the hand. Possible approach points are: 1) the highest location on top of the object with the hand coming from above (e.g. when grasping an object from above or placing it down); 2) center of the object with the hand current direction along the current $x$ and $y$ as computed from the current velocity vector (e.g., when grasping from the side); 3) the object center and the direction along a straight line from the hand to the object. These points are used to compute the most likely trajectory to the object given the current hand pose and velocity. 
At any time step, a Hobby spline \cite{Hobby1986} is computed to approximate the trajectory between the hand with its current direction and any of these approach points. These splines fit Bezier curves passing through the given start and end points and following the given tangent direction at these points. $n$ control points are then computed by the algorithm, along with covariances growing backward from the destination point. Here, to keep the computation time steady, $n$ = 8 for each trajectory was used, independently of the distance to the object\footnote{Considering a reaching distance $<1.5\ m$, this resolution is high enough to approximate the arm trajectories well.}.
From the current hand position and velocity, the hand trajectory is predicted via Kalman filtering with the same number of points as used for the splines, and with increasing variance for each further point, since the uncertainty about future positions of the hand grows with each time step. For each point $\pmb{\mu}_i$ in the predicted hand trajectory and in the estimated trajectories to the objects' approach points, hence a distribution $\mathcal{N}(\pmb{\mu}_i,\pmb{\Sigma}_i)$ can be defined. A measure of distance between two trajectories can thus be computed considering the weighted sum of the Bhattacharyya distances \cite{Kailath1967} between corresponding points on the trajectories. For each object, the approach point trajectory with closest distance to the current trajectory is chosen. To obtain features that express the likelihood of each object to be the target of the current movement, the sign of the sum of Bhattacharyya distances is changed and these are fed to a softmax function with temperature $0.05$. That is, for each object $o_i$ the corresponding TPA feature is computed as:
\begin{equation}
TPA(o_i) = \pmb{\sigma}(-\pmb{z}_i)
\end{equation}
where
\begin{equation}
\pmb{z}_i = \min\limits_{p\in \mathcal{AP}_{i}}(\sum_{j=0}^{n}w_j\cdot D_B[\mathcal{N}(\pmb{\mu_h}_j,\pmb{\Sigma_h}_j),\mathcal{N}(\pmb{\mu_{p}}_j,\pmb{\Sigma_{p}}_j)]) \ .
\end{equation}
Here $\pmb{\sigma}$ denotes the softmax function, $D_B[p(x),q(x)]$ the Bhattacharyya distance between two distributions, and $\mathcal{AP}_{i}$ the set of approach points for $o_i$. The weight $w_j$ is a function of the cosine between the normalized velocity vectors at point $j$ on the predicted hand trajectory and on the trajectory to object $i$, so the larger the difference in direction at the two points the larger the distance between those distributions is weighted. 
\begin{figure}[t]
    \centering
\includegraphics[width=0.47\textwidth]{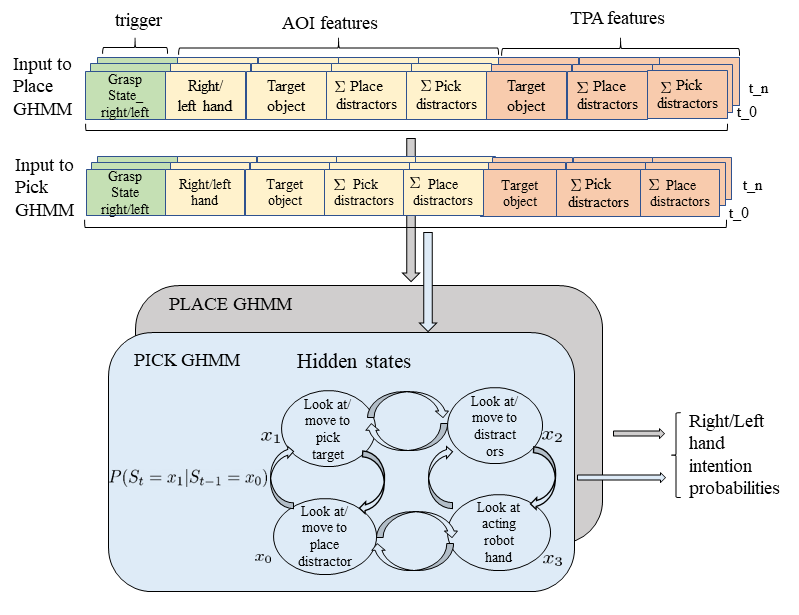}
\caption{Schematic representation of the Gaussian Hidden Markov Models used for intention estimation along with the rearranged features. The four hidden states are fully connected, here only connections between adjacent states are shown for readability. The place GHMM has a similar structure as the pick GHMM, but in this case place targets are considered.}
\label{fig:model}
\end{figure}
\section{GAUSSIAN HIDDEN MARKOV MODELS FOR INTENTION ESTIMATION}
\label{sec:model}
The inference regarding each one-handed action is done by two Gaussian Hidden Markov Models (GHMM), trained on pick and place sequences, respectively. Pick sequences are defined from the start of each trial until pressing of the grasping trigger and labeled with the corresponding pick target, while place sequences were defined from the pressing to the releasing of the trigger and labeled with the corresponding place target. This yields a multiclass classification problem with 12 classes (6 pick and 6 place intentions).

Each of the observation vectors in a sequence consisted of 26 features: 1 binary grasping state, 13 AOI features (6 pick, 6 place targets, the acting robot hand), 12 TPA features (same as for the AOIs but no robot hand)\footnote{Even if the other hand was in the scene, this in not used and never looked at for picking or placing with one hand. We consider here a system capable to work with just one hand first, addressing the extension to a bimanual scenario in the next section.}. AOI and TPA features were normalized to express the probability that each object is the current target of the gaze $\mathbf{g}_t$ and of the hand motion $\mathbf{h}_t$, i.e., $\mathrm{P}(\mathrm{AOI}=o_i|\mathbf{g}_t)$ and $\mathrm{P}(\mathrm{TPA}=o_i|\mathbf{h}_t)$, respectively. Still, this is a large number of features, which would change as soon as an object disappears or appears, or a new scene is presented. In order to make the approach scalable to other scenes than those seen in training and to limit the input to the model to semantically relevant cues describing the current sensorimotor state in a compact form, features were rearranged in the following form to entail just 8 features:
\begin{multline}
    \mathbf{F_{pick}}_t = \{grasping\_state_t,\mathrm{P}(\mathrm{AOI}=a_i|\mathbf{g}_t),\\
    \sum_{j\neq i}\mathrm{P}(\mathrm{AOI}=a_j|\mathbf{g}_t),  \sum_{k=1}^{m}\mathrm{P}(\mathrm{AOI}=b_k|\mathbf{g}_t),\\
    \mathrm{P}(\mathrm{AOI}=Robot|\mathbf{g}_t), 
    \mathrm{P}(\mathrm{TPA}=a_i|\mathbf{h}_t),\\
    \sum_{j\neq i}\mathrm{P}(\mathrm{TPA}=a_j|\mathbf{h}_t),  \sum_{k=1}^{m}\mathrm{P}(\mathrm{TPA}=b_k|\mathbf{h}_t)\} \quad .
\end{multline}

This represents the input vector for the 'pick' GHMM, while for the 'place' GHMM a similarly arranged vector is obtained by replacing in the above equation the $a$ terms (picking targets) with the $b$ and viceversa. For the place rearrangement, the TPA feature of the object in hand was forced to be slightly lower than the furthest object, so that it would not dominate the sum of distractors and the place target TPA feature would increase steadily as the hand approaches it.\par
These rearrangements allow us on the one hand to train just one model for the 'pick' and one for the 'place' intention, and on the other hand to have a system independent of the number of objects in the scene. The multiclass classification problem is then tackled in a one-vs-all fashion.
Each GHMM is defined with 4 states that might represent moving/looking toward the action target or the distractors in either action, or looking at the own moving hand.

Training is done by rearranging the sequences according to the current ground truth target (e.g. for "pick at $a_0$", by considering the AOI/TPA for $a_0$ as single features and summing other $a_i$ and $b_j$ together in the other two features) and feeding each GHMM the corresponding sequences. Testing and online prediction is done considering a window of $\Delta t = 0.45s$ and rearranging features in this window according to each possible 'pickable' or 'placeable' target candidate and computing as many likelihood scores (i.e., 12, in our scene). The score with highest log-probability and exceeding a threshold of 0 is taken as prediction of the respective intention. Otherwise, the confidence in the prediction is not large enough and no prediction is output. Exploration of the predicted log-likelihood scores indeed show that these are distributed bimodally, with very negative values when no intention is confidently recognized and sudden positive values as soon as an action is recognized (see Fig. \ref{fig:time_evol}).  Ofﬂine training and online recognition are implemented in Python by means of the
hmmlearn library \footnote{\url{https://hmmlearn.readthedocs.io}}. The model with its input is sketched in Fig. \ref{fig:model}.

\section{Bimanual architecture}\label{sec:bim}
Upon assessment of generalization of the models across hands, we investigated a bimanual architecture considering the combination of intentions coming from two effectors.
Both the AOIs and TPAs features as formulated here are not adapted for a specific effector, thus we reasoned that, in principle, each model could be used interchangeably for the left or right hand.  This means also that the same model could be instantiated for each hand: each instantiation receives the same gaze data (up to the 'robot hand' AOI) but different grasping states and TPA features. 
In this way, concurrent but independent picking and placing intentions for both hands can be recognized or more coordinated bimanual actions can be considered.
This can be done by learning how the two outputs from the one-hand models relate to labeled one-hand or bimanual intentions but in a simple pick-and-place scenario a rule-based system can suffice.
In this way, a hand-over or other more complex manipulations can be broken down in the intentions assigned to each effector. 

Such a rule-based system was implemented to handle some basic bimanual tasks by the following rules: 1) if for both hands 'pick object x' is recognized then a bimanual pick is recognized ('Multihand Pick'); 2) if holding and placing down the same object is recognized for both hands then a bimanual place is recognized at bimanual level ('Multihand Place'); 3) by considering each hand as a possible target surface, if for one hand a placing to the other hand is recognized and for the other hand picking the object is recognized, then a handover from the first hand to the second is output as bimanual intention ('Hand Over').
As mentioned, these rules combine the information provided by the intention modules from each hand, i.e., the predicted one-hand intention for each hand ('Pick' or 'Place'), and additionally the target object of each hand.

\section{RESULTS}
\label{sec:results}

\begin{figure}[t]
    \centering
\includegraphics[width=0.47\textwidth]{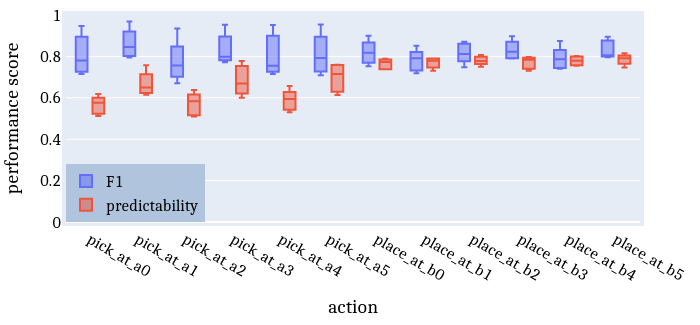}
\caption{F1 and predictability scores for each intention across 5-fold cross-validation over 10 participants. Here, predictability represents the portion of observations for which a confident prediction is available.}
\label{fig:F1_pred}
\end{figure} 
\begin{figure}[t]
    \centering
\includegraphics[width=0.47\textwidth]{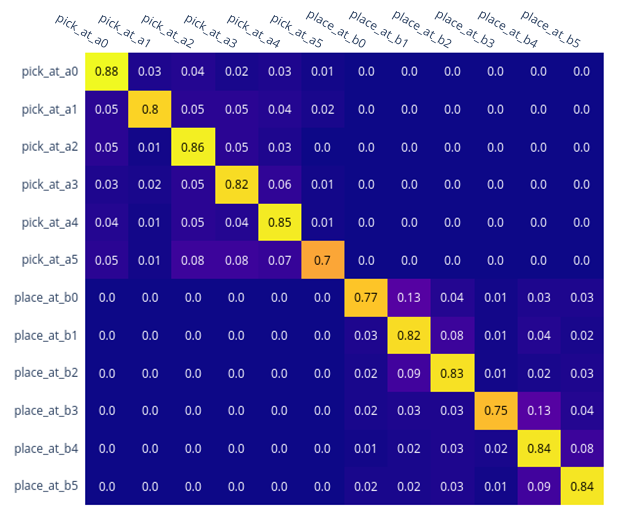}
\caption{Normalized confusion matrix: on the $x$ axis the predictions, on the $y$ the ground truth.  Numbers represents the frequency with which samples of a certain class (row) were classified with the label on the corresponding column. Some confusion is present between neighboring targets/location, but no confusion between pick and place intentions.}
\label{fig:cm}
\end{figure}

\subsection{Cross-validation across users}
To assess generalizability across users  a 5-fold cross-validation was conducted, by training each time on sequences by 8 participants and testing on sequences by 2 unseen participants. Training was conducted for the one-handed blocks considering the TPA features from the acting hand. For the hand-over blocks, we considered the picking hand for the picking sequences and the placing hand for the placing sequences.
F1 scores across folds for each action are summarized in the boxplot in Fig.~\ref{fig:F1_pred}. Here, it must be noted that these scores are computed considering only samples for which a prediction is indeed available (i.e., log-likelihood score over 0 for at least one of the classes). It can be seen that for each intention the F1 score, which combines precision and recall, is over 70\%. Yet, to put this information in context, in the same plot also the predictability is shown, defined as the ratio of each action for which a prediction is available, independently of its accuracy. For each action, on average at least 60\% of the time an intention is recognized with sufficient confidence. This relates to earliness of prediction, considering the complementary view that a prediction is first available after 40\% of the action has unfolded. Moreover, place intentions appear to be recognized earlier and better than pick intentions: the place locations indeed have less distractors in the vicinity and are directly targeted by the gaze, while to pick a cylinder in the two-lines configuration the eyes check first other cylinders close by to prevent the hand from colliding with them.
Looking at the confusion matrix in Fig.~\ref{fig:cm}, it is apparent that wrong predictions happen mostly between adjacent target locations, but the two general intentions (pick and place) are never confused. As in \cite{Fuchs2021}, this happens because the models learn that the grasping state is the most predictive feature for the type of intention.

\begin{figure}[t]
    \centering
\includegraphics[width=0.40\textwidth]{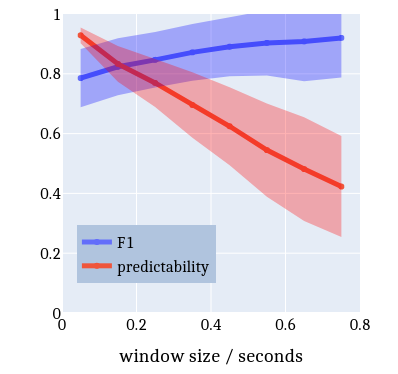}
\caption{F1 and predictability as functions of the window size for the sequence in input to the GHMMs. The longer the sequence observed the better the prediction, yet this becomes available only toward the end of the action, causing the predictability (portion of observations with a confident prediction) to drop dramatically.}
\label{fig:wind_size}
\end{figure}

\subsection{Earliness of prediction}
The dependency of the performance and predictability on the window size is further shown in Fig. \ref{fig:wind_size}. We chose $\Delta t$= 0.45\,s since for that value the overall predictability is above 60\% and the overall F1 score above 80\%.

\begin{figure}[t]
    \centering
\includegraphics[width=0.47\textwidth]{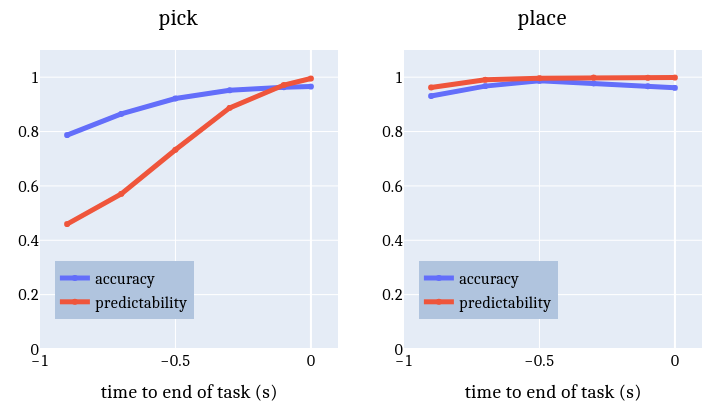}
\caption{Evolution in time of the accuracy and predictability from 900 ms until the end of the task (right: the pick task, left: the place task).}
\label{fig:acc_evol}
\end{figure}

\begin{figure*}[t]
    \centering
\includegraphics[width=0.89\textwidth]{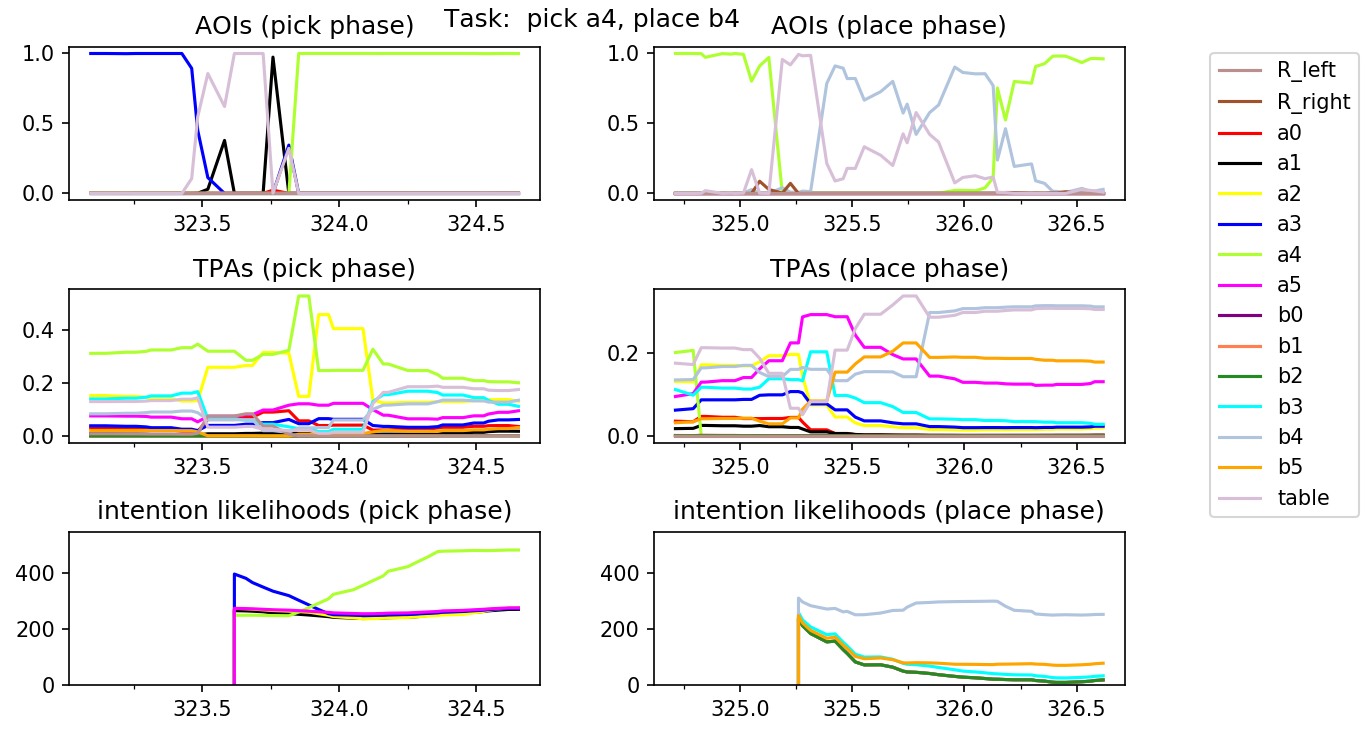}
\caption{Evolution in time of the AOI and TPA features and of the intention log-likelihoods during an exemplary pick (on the left) and place (on the right) task with the right hand. The intention log-likelihoods are plotted before normalization to appreciate the difference between different intentions. In these plots, also the AOI and TPA likelihoods for the table and the non-acting hand are shown for reference, but they were not considered as input to the GHMMs.}
\label{fig:time_evol}
\end{figure*}

The earliness of prediction is indeed a critical feature, since once an intention is recognized confidently, the robot needs time to plan an appropriate assisting action. To get a feeling about the absolute times when first a prediction is available, instead of a relative portion of the action, accuracy and predictability were computed over all trials at 100\,ms intervals before the end of the task. These were computed separately for pick and place intentions, since they strongly differ in their predictability, as shown in Fig.~\ref{fig:F1_pred}. Median execution times were between 1.1\,s and 1.5\,s for pick intentions and between 2.2 and 2.6 s for place intentions. As it can be noted in Fig.~\ref{fig:acc_evol}, left, for pick intentions accuracy is over 80\% between -900 ms and 0 ms, but predictability steeply increases as the model accumulates evidence for the target over distractors. In the last half second, predictability is then above 70\%. In natural eye-hand coordination,  500 ms ahead is also typically the time when people consistently fixate on the object target of the manipulation before their hand reaches it  \cite{Johansson2001,Hayhoe2003}. For place intentions on the other hand, accuracy and predictability are constantly high with some lower accuracy in the beginning and in the end. In these two moments, indeed, the gaze is more focused on the grasped object, rather than on the place target: in the beginning because after grasping without haptic feedback the user makes sure that the object moves along with the hand by visually monitoring it; at the end because when placing down from above the object finally occludes the coaster. This behavior is exemplified on the top panel right in Fig.~\ref{fig:time_evol}, where it appears that the AOI likelihood of the grasped object is highest at the beginning and end of the placing phase.

\begin{table*}[t]
\caption{Accuracy and predictability for different combinations of features}
\label{tab:ablation}
\begin{center}
\begin{tabular}{|c||c|c|c|c|c|c|c|c|}
\hline
 \bf{Feature combination} & \multicolumn{2}{c|}{\bf{AOI + TPA + GS}} & \multicolumn{2}{c|}{\bf{AOI + GS}} & \multicolumn{2}{c|}{\bf{TPA + GS}} & \multicolumn{2}{c|}{\bf{AOI + TPA}}  \\
\hline
\underline{cross-validation} & \underline{users} & 
\underline{hands} & \underline{users} &
\underline{hands} & \underline{users} &
\underline{hands} & \underline{users} &
\underline{hands}\\
\hline
F1 & \bf{82\%} & 80\% & 80\% & 81\% & 38\% & 45\% & 48\% & 50\% \\
\hline
Predictability & 71\% & 73\% & 71\% & 73\% & 71\% & 73\% & \bf{99}\% & \bf{99}\% \\
\hline
\end{tabular}
\end{center}
\end{table*}

\subsection{Generalization across effectors and ablation study}
With the prospect of a bimanual scenario, we were also interested in assessing whether the models would work equally well when using the non-dominant hand, when perhaps movements are less fast and dexterous and eye movements could display a slightly different behavior.  Moreover, we also wanted to ascertain how reliable the different kinds of features are, and which are the most effective combinations. Hence, we conducted a kind of ablation analysis, training each time on a different combination of features and comparing it to the full-fledged system. 
The results of this analyses are presented in Table~\ref{tab:ablation}. For each combination the cross-validation was conducted across users (5-fold) and across hands (2-fold, training on trials performed with the right hand and testing on left hand trials, and vice versa). In general, the models generalize well across hands, that is,  even when training with 50\% of the trials, F1 scores are generally on par with those in the 5-fold cross-validation.\par 
The best performance across users is obtained when all features are considered (AOI, TPA and grasping state), although the improvement is small w.r.t. only considering gaze and grasping state (AOI + GS). This is the case because, as illustrated for an exemplary task in Fig.~\ref{fig:time_evol}, the gaze is faster than the hand, hence provides more punctual and directed cues, landing on the target object early on, both in the pick and place phase. The TPA feature of the target, on the other hand, slowly rises to be the winning one as the hand approaches the target. For most part of the action the hand trajectory is compatible with multiple objects in the cluttered scene. Still, TPA features provide some predictive information, considering the performance of TPA + GS across 12 possible intentions, and in scenarios where more complex intentions need to be recognized they might complement gaze predictivity earlier, for example by discriminating between two possible actions on the same object. 

Considering other systems, intention estimation frameworks for shared control are usually built on specific setups and datasets, making a direct comparison difficult. Results presented in \cite{Fuchs2021} showed that simply considering fixations on objects (i.e. the winning AOI in a certain time window, without GHMM) performs worse than the generative model approach, since this method deals worse with spurious glances to nearby objects. This is even more the case here where often distractors are glanced for obstacle avoidance. In  the ablation study still, the complete system shows an improvement with respect to our older model (AOI+GS), suggesting that the use of all features with the GHMM yields an even better performance.

\section{DISCUSSION AND CONCLUSIONS}\label{sec:conclusions}

In this study, an intention estimation system was presented, relying on gaze and hand motion features and using GHMMs for inference. The framework was trained on a dataset purposely collected in VR and containing both hands executing pick and place and handover tasks. The accuracy of the intention recognition stays high across different users and across hands, even in a rather cluttered scene, with occlusions and objects presented close together. A reliable prediction is available early on, before the movement is executed halfway and in general 0.5\,s before the end of the action, when a grasping or placing assistance could be provided by the robotic partner. The system works best when using all features together, yet the most decisive are the gaze and the grasping state. Still, motion features are relevant in distinguishing one-hand intentions during bimanual actions. The simulated scenario presented here is indeed intended for bimanual manipulations and data were collected for both hands in each block. 
A simple rule-based system was proposed in the described setup and with simple bimanual intentions, but it is prone to the stability and correctness of the learned one-hand intention modules.
More complex bimanual manipulations and the learning of an explicit bimanual intention module are part of current and future work, where the rule-based system will provide a basic baseline in addition.

Moreover, although hands can act in parallel, the eyes can subserve the information and supervising needs of only one effector at a time. This is an open problem since most current gaze-based systems, to our knowledge, deal with the control of just one arm, while most recent bimanual frameworks in shared control rely on learning directly both hand motion patterns (as in \cite{Rakita2019}) or the second robotic arm is autonomously coordinated with the motion of the operator's right arm \cite{Laghi2018}. In natural bimanual eye-hand coordination, we usually visually monitor more closely the effector executing the most difficult action and rely on covert attention, proprioception and haptic feedback to control the other limb \cite{Srinivasan2010}. Alternatively, we shift our eyes between the two targets in an interleaved fashion \cite{Riek2003}. Current and future work is therefore focused on investigating how further actions can be introduced in our framework and how with the same gaze features and different hand features independent and coordinated bimanual actions can be effectively recognized.

\addtolength{\textheight}{-1cm}   % This command serves to balance the column lengths

%%%%%%%%%%%%%%%%%%%%%%%%%%%%%%%%%%%%%%%%%%%%%%%%%%%%%%%%%%%%%%%%%%%%%%%%%%%%%%%%

%%%%%%%%%%%%%%%%%%%%%%%%%%%%%%%%%%%%%%%%%%%%%%%%%%%%%%%%%%%%%%%%%%%%%%%%%%%%%%%%

%%%%%%%%%%%%%%%%%%%%%%%%%%%%%%%%%%%%%%%%%%%%%%%%%%%%%%%%%%%%%%%%%%%%%%%%%%%%%%%%
%\section*{APPENDIX}


\section*{ACKNOWLEDGMENT}
 We acknowledge and thank for the collaboration and the fruitful discussions the colleagues Soshi Iba, Karankumar Patel, Nanami Tsukamoto, and Naoki Hosomi. We further thank Niels Einecke for helping with the AOI occlusion handling.

%%%%%%%%%%%%%%%%%%%%%%%%%%%%%%%%%%%%%%%%%%%%%%%%%%%%%%%%%%%%%%%%%%%%%%%%%%%%%%%%

\bibliographystyle{IEEETran.bst} 
\bibliography{iros.bib}

\begin{thebibliography}{10}
\providecommand{\url}[1]{#1}
\csname url@rmstyle\endcsname
\providecommand{\newblock}{\relax}
\providecommand{\bibinfo}[2]{#2}
\providecommand\BIBentrySTDinterwordspacing{\spaceskip=0pt\relax}
\providecommand\BIBentryALTinterwordstretchfactor{4}
\providecommand\BIBentryALTinterwordspacing{\spaceskip=\fontdimen2\font plus
\BIBentryALTinterwordstretchfactor\fontdimen3\font minus
  \fontdimen4\font\relax}
\providecommand\BIBforeignlanguage[2]{{%
\expandafter\ifx\csname l@#1\endcsname\relax
\typeout{** WARNING: IEEEtran.bst: No hyphenation pattern has been}%
\typeout{** loaded for the language `#1'. Using the pattern for}%
\typeout{** the default language instead.}%
\else
\language=\csname l@#1\endcsname
\fi
#2}}

\bibitem{Choi2018}
P.~J. Choi, R.~J. Oskouian, and R.~S. Tubbs, ``Telesurgery: past, present, and
  future,'' \emph{Cureus}, vol.~10, no.~5, 2018.

\bibitem{Herlant2016}
L.~V. Herlant, R.~M. Holladay, and S.~S. Srinivasa, ``Assistive teleoperation
  of robot arms via automatic time-optimal mode switching,'' in \emph{2016 11th
  ACM/IEEE International Conference on Human-Robot Interaction (HRI)}.\hskip
  1em plus 0.5em minus 0.4em\relax IEEE, 2016, pp. 35--42.

\bibitem{Artigas2016}
J.~Artigas, R.~Balachandran, C.~Riecke, M.~Stelzer, B.~Weber, J.-H. Ryu, and
  A.~Albu-Schaeffer, ``Kontur-2: force-feedback teleoperation from the
  international space station,'' in \emph{2016 IEEE International Conference on
  Robotics and Automation (ICRA)}.\hskip 1em plus 0.5em minus 0.4em\relax IEEE,
  2016, pp. 1166--1173.

\bibitem{Petereit2019}
J.~Petereit, J.~Beyerer, T.~Asfour, S.~Gentes, B.~Hein, U.~D. Hanebeck,
  F.~Kirchner, R.~Dillmann, H.~H. G{\"o}tting, M.~Weiser, \emph{et~al.},
  ``Robdekon: Robotic systems for decontamination in hazardous environments,''
  in \emph{2019 IEEE International Symposium on Safety, Security, and Rescue
  Robotics (SSRR)}.\hskip 1em plus 0.5em minus 0.4em\relax IEEE, 2019, pp.
  249--255.

\bibitem{Griffin2005}
W.~B. Griffin, W.~R. Provancher, and M.~R. Cutkosky, ``Feedback strategies for
  telemanipulation with shared control of object handling forces,''
  \emph{Presence}, vol.~14, no.~6, pp. 720--731, 2005.

\bibitem{Rakita2019}
D.~Rakita, B.~Mutlu, M.~Gleicher, and L.~M. Hiatt, ``Shared control--based
  bimanual robot manipulation,'' \emph{Science Robotics}, vol.~4, no.~30, 2019.

\bibitem{Fuchs2021}
S.~Fuchs and A.~Belardinelli, ``Gaze-based intention estimation for shared
  autonomy in pick-and-place tasks,'' \emph{Frontiers in Neurorobotics},
  vol.~15, p.~33, 2021.

\bibitem{Hauser2013}
K.~Hauser, ``Recognition, prediction, and planning for assisted teleoperation
  of freeform tasks,'' \emph{Autonomous Robots}, vol.~35, no.~4, pp. 241--254,
  2013.

\bibitem{Dragan2013}
A.~D. Dragan and S.~S. Srinivasa, ``A policy-blending formalism for shared
  control,'' \emph{The International Journal of Robotics Research}, vol.~32,
  no.~7, pp. 790--805, 2013.

\bibitem{Javdani2015}
S.~Javdani, S.~S. Srinivasa, and J.~A. Bagnell, ``Shared autonomy via hindsight
  optimization,'' in \emph{Robotics science and systems: online proceedings},
  vol. 2015.\hskip 1em plus 0.5em minus 0.4em\relax NIH Public Access, 2015.

\bibitem{Aarno08}
D.~Aarno and D.~Kragic, ``Motion intention recognition in robot assisted
  applications,'' \emph{Robotics and Autonomous Systems}, vol.~56, no.~8, pp.
  692--705, 2008.

\bibitem{Tanwani2017}
A.~K. Tanwani and S.~Calinon, ``A generative model for intention recognition
  and manipulation assistance in teleoperation,'' in \emph{2017 IEEE/RSJ
  International Conference on Intelligent Robots and Systems (IROS)}.\hskip 1em
  plus 0.5em minus 0.4em\relax IEEE, 2017, pp. 43--50.

\bibitem{Hayhoe2003}
M.~M. Hayhoe, A.~Shrivastava, R.~Mruczek, and J.~B. Pelz, ``Visual memory and
  motor planning in a natural task,'' \emph{Journal of vision}, vol.~3, no.~1,
  pp. 6--6, 2003.

\bibitem{Johansson2001}
R.~S. Johansson, G.~Westling, A.~B{\"a}ckstr{\"o}m, and J.~R. Flanagan,
  ``Eye--hand coordination in object manipulation,'' \emph{Journal of
  Neuroscience}, vol.~21, no.~17, pp. 6917--6932, 2001.

\bibitem{Belardinelli2016}
A.~Belardinelli, M.~Y. Stepper, and M.~V. Butz, ``It's in the eyes: Planning
  precise manual actions before execution,'' \emph{Journal of vision}, vol.~16,
  no.~1, pp. 18--18, 2016.

\bibitem{Shafti2019}
A.~Shafti, P.~Orlov, and A.~A. Faisal, ``Gaze-based, context-aware robotic
  system for assisted reaching and grasping,'' in \emph{2019 International
  Conference on Robotics and Automation (ICRA)}.\hskip 1em plus 0.5em minus
  0.4em\relax IEEE, 2019, pp. 863--869.

\bibitem{Aronson2021}
R.~M. Aronson, N.~Almutlak, and H.~Admoni, ``Inferring goals with gaze during
  teleoperated manipulation,'' in \emph{2021 IEEE/RSJ International Conference
  on Intelligent Robots and Systems (IROS)}.\hskip 1em plus 0.5em minus
  0.4em\relax IEEE, pp. 7307--7314.

\bibitem{Fathaliyan2018}
A.~Haji~Fathaliyan, X.~Wang, and V.~J. Santos, ``Exploiting three-dimensional
  gaze tracking for action recognition during bimanual manipulation to enhance
  human--robot collaboration,'' \emph{Frontiers in Robotics and AI}, vol.~5,
  p.~25, 2018.

\bibitem{Wang2020}
X.~Wang, A.~Haji~Fathaliyan, and V.~J. Santos, ``Toward shared autonomy control
  schemes for human-robot systems: Action primitive recognition using eye gaze
  features,'' \emph{Frontiers in Neurorobotics}, vol.~14, p.~66, 2020.

\bibitem{Admoni2016}
H.~Admoni and S.~Srinivasa, ``Predicting user intent through eye gaze for
  shared autonomy,'' in \emph{2016 AAAI Fall Symposium Series}, 2016.

\bibitem{Jain2019}
S.~Jain and B.~Argall, ``Probabilistic human intent recognition for shared
  autonomy in assistive robotics,'' \emph{ACM Transactions on Human-Robot
  Interaction (THRI)}, vol.~9, no.~1, pp. 1--23, 2019.

\bibitem{Razin2017}
Y.~Razin and K.~Feigh, ``Learning to predict intent from gaze during robotic
  hand-eye coordination,'' in \emph{Thirty-First AAAI Conference on Artificial
  Intelligence}, 2017.

\bibitem{Li2020}
S.~Li, M.~Bowman, H.~Nobarani, and X.~Zhang, ``Inference of manipulation intent
  in teleoperation for robotic assistance,'' \emph{Journal of Intelligent \&
  Robotic Systems}, pp. 1--15, 2020.

\bibitem{Herbort2011}
O.~Herbort and M.~V. Butz, ``Habitual and goal-directed factors in (everyday)
  object handling,'' \emph{Experimental Brain Research}, vol. 213, no.~4, pp.
  371--382, 2011.

\bibitem{Strasburger2011}
H.~Strasburger, I.~Rentschler, and M.~J{\"u}ttner, ``Peripheral vision and
  pattern recognition: A review,'' \emph{Journal of vision}, vol.~11, no.~5,
  pp. 13--13, 2011.

\bibitem{Muehlig2014}
M.~Mühlig, A.~Hayashi, M.~Gienger, S.~Iba, and T.~Yoshiike, ``Receding horizon
  optimization of robot motions generated by hierarchical movement
  primitives,'' in \emph{2014 IEEE/RSJ International Conference on Intelligent
  Robots and Systems}, 2014, pp. 129--135.

\bibitem{Iba2019}
\BIBentryALTinterwordspacing
S.~Iba, ``State prediction system,'' U.S. Patent 10\,166\,677, Jan.~1, 2019.
  [Online]. Available: \url{https://patents.google.com/patent/US10166677}
\BIBentrySTDinterwordspacing

\bibitem{Hobby1986}
J.~D. Hobby, ``Smooth, easy to compute interpolating splines,'' \emph{Discrete
  \& Computational Geometry}, vol.~1, no.~2, pp. 123--140, 1986.

\bibitem{Kailath1967}
T.~Kailath, ``The divergence and {B}hattacharyya distance measures in signal
  selection,'' \emph{IEEE transactions on communication technology}, vol.~15,
  no.~1, pp. 52--60, 1967.

\bibitem{Laghi2018}
M.~Laghi, M.~Maimeri, M.~Marchand, C.~Leparoux, M.~Catalano, A.~Ajoudani, and
  A.~Bicchi, ``Shared-autonomy control for intuitive bimanual
  tele-manipulation,'' in \emph{2018 IEEE-RAS 18th International Conference on
  Humanoid Robots (Humanoids)}.\hskip 1em plus 0.5em minus 0.4em\relax IEEE,
  2018, pp. 1--9.

\bibitem{Srinivasan2010}
D.~Srinivasan and B.~J. Martin, ``Eye--hand coordination of symmetric bimanual
  reaching tasks: Temporal aspects,'' \emph{Experimental Brain Research}, vol.
  203, no.~2, pp. 391--405, 2010.

\bibitem{Riek2003}
S.~Riek, J.~R. Tresilian, M.~Mon-Williams, V.~L. Coppard, and R.~G. Carson,
  ``Bimanual aiming and overt attention: one law for two hands,''
  \emph{Experimental brain research}, vol. 153, no.~1, pp. 59--75, 2003.

\end{thebibliography}

\end{document}